\documentclass[11pt,a4paper]{article}
\usepackage[hyperref]{acl2021}
\aclfinalcopy  
\usepackage{times}
\usepackage{latexsym}

\usepackage{graphicx}
\usepackage{amsfonts}
\usepackage{xcolor}         

\graphicspath{{Graphs/}} 
\usepackage{mathtools, cuted}
\usepackage{scalerel,graphicx,xparse}

\NewDocumentCommand\emojione{}{\scalerel*{\includegraphics{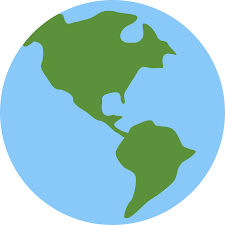}}{X}}

\usepackage{microtype}


\title{More than Correlation: \\Do Large Language Models Learn Causal Representations of Space? \emojione{}{}}

\author{Yida Chen\footnotemark[1] ,~ Yixian Gan\footnotemark[1] ,~ Sijia Li\footnotemark[1] ,~ Li Yao\footnotemark[1] ,~ Xiaohan Zhao\thanks{Names are listed in alphabetical order of the last names not by contribution. Correspondence should go to \{\href{mailto:yidachen@g.harvard.edu}{yidachen}, \href{mailto:ygan@g.harvard.edu}{ygan}, \href{mailto:sijiali@g.harvard.edu}{sijiali}, \href{mailto:liyao@g.harvard.edu}{liyao}, \href{mailto:xiaohanzhao@g.harvard.edu}{xiaohanzhan}\}@g.harvard.edu} \\
        Harvard University\\ Cambridge, MA 02138\\ \{\href{mailto:yidachen@g.harvard.edu}{yidachen}, \href{mailto:ygan@g.harvard.edu}{ygan}, \href{mailto:sijiali@g.harvard.edu}{sijiali}, \href{mailto:liyao@g.harvard.edu}{liyao}, \href{mailto:xiaohanzhan@g.harvard.edu}{xiaohanzhan}\}@g.harvard.edu }

\begin{document}
\maketitle
\begin{abstract}
Recent work found high mutual information between the learned representations of large language models (LLMs) and the geospatial property of its input, hinting an emergent internal model of space. However, whether this internal space model has any causal effects on the LLMs' behaviors was not answered by that work, led to criticism of these findings as mere statistical correlation. Our study focused on uncovering the causality of the spatial representations in LLMs. In particular, we discovered the potential spatial representations in DeBERTa, GPT-Neo using representational similarity analysis and linear and non-linear probing. Our casual intervention experiments showed that the spatial representations influenced the model's performance on next word prediction and a downstream task that relies on geospatial information. Our experiments suggested that the LLMs learn and use an internal model of space in solving geospatial related tasks.

\end{abstract}

\section{Introduction}
Whether language models develop an internal understanding of the world they interact with remains a contentious topic. Some research suggests that language models inherently possess world knowledge even without explicit grounding~\cite{petroni2019language, abdou2021can, li2021implicit}, other works argue that the generative neural networks are merely aggregating the surface statistics they observed in the training data~\cite{bender2020climbing,bisk2020experience, merrill2021provable}. A recent article~\cite{gurnee2023language} claims that the large language model (LLM), LLaMa~\cite{touvron2023llama}, has an internal world model of space by showing that LLM's intermediate activation of a location word can be map linearly to its physical location. However, this research did not establish a causal link between this internal spatial representations and the model's behaviors. As argued by~\cite{lecun2022path}, the world model enables agents, both natural and artificial ones, to predict the consequence of their action so they can better plan and explore new solutions to the new problem. Whether the world model of space guides the actions of the LLM was missing in the discussion of ~\cite{gurnee2023language}. To fill in this blank, our work investigated whether editing the language model's internal representations of space can affect its output accordingly --- that is, if there exists a causal relation between the world model of space and the LLM's actions. 

We used DeBERTa-v2 (1.5B)~\cite{he2020deberta} and GPT-Neo (1.3B)~\cite{gpt-neo} in our experiments. Using representational similarity analysis and probing classifiers, we showed that DeBERTa-v2 and GPT-Neo have implicitly learned spatial representations through pretraining. In particular, our experiments showed that the distance between the hidden states of city names in the DeBERTa's learned representation space has high correlation with the real-world geospatial distance between cities. Through linear and non-linear probing regressors, we further demonstrated that the hidden states of city names can be mapped linearly and non-linearly (with higher accuracy) to the cities' actual latitudes and longitudes. 

Furthermore, our research established a causal connection between the internal spatial representations of LLMs and their performances. Given a classifier that predicts the country of a city using the LLM's last layer hidden states, we can improve the classifier's accuracy by modifying the intermediate hidden states of the input city names using gradients of the probing classifier so that the internal spatial representations are more precisely mapped to the cities' real geographical locations. Conversely, accuracy decreases when we forced the alignment between the LLM's internal spatial representations and actual geospatial location to be less precise. We also conducted an experiment to examine whether internal spatial representations play a causal role in the language model's ability to predict the next word that contains geospatial information, such as the associated country of a city. We observed an improvement in the accuracy of predicting next word by perturbing hidden representations with enhanced geospatial information, while we observed deterioration by intervention to force less precise geospatial information. These results showed that LLMs learn an internal world model of space that governs their behavior, beyond simply memorizing the training data.

\section{Related Work}

Recent work~\cite{gurnee2023language} found linear representations of space and time in LLaMa~\cite{touvron2023llama} using probing classifiers. This approach involves creating a dataset of internal representations of texts or words. In the context of transformer-based language models, these internal representations typically refer to the intermediate activation layers of the input token across various transformer layers. To analyze whether these representations encode meaningful property of input, previous works trained an auxiliary classifier~\cite{alain2016understanding} that mapped the internal representations to a specific property of their corresponding text. For the investigation on spatial and temporal representations, the property could be the two-dimensional latitude and longitude coordinates. High prediction performance indicates a strong mutual information between the LLM's learned representations of the text and the spatial and temporal information embodied in the text.
\begin{table*}[t]
    \centering
    \begin{tabular}{||c|c|c|c|c|c||}
    \hline\hline
    city&city\_ascii&lat&lng&admin\_name&country \\ \hline
    Tokyo&Tokyo	&35.6897&139.6922&Tōkyō&Japan\\
    New York&New York&40.6943&-73.9249&New York&United States\\
    Shanghai&Shanghai&31.1667&121.4667&Shanghai&China\\
    São Paulo&Sao Paulo&-23.5500&-46.6333&São Paulo&Brazil\\\hline
    \end{tabular}
    \caption{Sample city names in the dataset}
    \label{tab:sample_cityname}
\end{table*}
\begin{table*}[t]
    \centering
    \begin{tabular}{||c|c|c|c|c|c|c||}
    \hline\hline
    city&city\_ascii&lat&lng&admin\_name&country&city\_concat\\ \hline
    London&London&51.5072&-0.1275&London, City of&United Kingdom&London, London, City of\\
    London&London&42.9836&-81.2497&Ontario&Canada&London, Ontario\\\hline
    \end{tabular}
    \caption{City names after concatenation. Since certain cities has duplicated names. To avoid the ambiguity, we concatenated the city's name with its administrative district.}
    \label{tab:sample_cityname_concat}
\end{table*}

To uncover the \textit{causal} role of learned representations in language models, previous studies intervened the internal representations of language models and observed the changes in the models' outputs. For example, \cite{giulianelli2018under} and \cite{tucker2021if} used the gradients of a probe $g_\theta$ to update the model representations $f(x)$ to measure the effect on the original model performance. Another approach involves intervening the representations by removing the property of interest $z$ from $f(x)$ to understand the causal effects of $z$ on model performance. This can be achieved by repeatedly training a probe $g_\theta$ and projecting the property out of the representations \cite{elazar2021amnesic}, or by training $g_{\theta}$ adversarially \cite{feder2021causalm}. In addition, \cite{hernandez2023linearity} assessed the causality of a linear relational embedding (LRE) on the language model by inverting LRE on an object from a different subject-object pair to perturb the representations of a given subject.

As for the transformer-based language models, the BERT family models such as ~\cite{devlin2018bert}, ~\cite{sanh2019distilbert}, and ~\cite{he2020deberta} are specifically designed to learn rich word embeddings and have been performing well in a variety of downstream tasks in the past few years. Recent research showed that the GPT family models ~\cite{radford2018improving, radford2019language, brown2020language, openai2023gpt4, gpt-neo} and the LLaMa family models ~\cite{touvron2023llama} also learn the world model ~\cite{abdou2021can, gurnee2023language}. Understanding whether or not there exists a causal relation between the world model and various LLMs' hidden representations contributes to improved interpretability of LLM.

\section{Experimental Setup}

\subsection{Dataset and Preprocessing}
To examine the spatial representations within the language models, we used the world cities dataset \footnote{https://simplemaps.com/data/world-cities}, comprising city names, administrative districts, countries, and their respective longitudes and latitudes. We kept the top 100 countries with the highest city count. Table \ref{tab:sample_cityname} shows a few example of city names in the dataset. This dataset provides city names both in their original language and in ASCII characters. We used the name of cities in their original language in all of our experiments. To mitigate the issue of duplicate city names, we combined each city name with its administrative district, as illustrated in Table~\ref{tab:sample_cityname_concat}. 

\subsection{Prompt}
Besides geographical location, a city's name may imply other information such as demographics, economic growth, etc. In order to extract LLM's spatial information exclusively from city names, we used the following prompt to retrieve the learned representations:
\begin{itemize}
    \item[] Prompt: \texttt{The latitude and longit-} \texttt{ude of <name of the city> is}
\end{itemize}
All inputs are padded to the same length, and the hidden states of all tokens after each hidden layer, followed by a MeanPooling operation, are extracted as the representation of the city. 

\subsection{Models}
Previous work primarily focused on the representations from multibillion-parameter LLMs like the LLaMa family~\cite{gurnee2023language}. However, we wondered if the spatial representations have already been learned in smaller language models, provided that some works have used their fine-tuned versions to solve spatial tasks~\cite{shin2020bert, dan2020understanding}. In this work, we extracted representations of city names from two relatively small language models, DeBERTa-v2 and GPT-Neo, both of which have less than 2 billion parameters.

\section{Representational Similarity Analysis (RSA)}
RSA identifies the relations between two different representations of the same data. It was first proposed by ~\cite{Kriegeskorte2008} in neuroscientific studies. \cite{abdou2021can} applied this technique to measure the alignment of colors in the perceptual CIELAB color space with their textual representations learned by DeBERTa-v2. 

In our experiment, we employed this approach to study the correlation between two different spatial representations of countries: (1) the real geographical representations of countries in terms of longitude and latitude (averaged longitude and latitude of cities in each country), and (2) the LLM activation representations of countries (the 16th layer for DeBERTa-v2 and the 24th layer for GPT-Neo). We computed the distance between each pair of countries for both representations, obtaining two $N \times N$ matrices ($N$ is the number of countries). For the longitude-latitude representations, the real geographical distances~\cite{Karney_2010} between the two countries are computed. For the LLM activation representations, three different distance metrics are used for each country pair, namely Spearman correlation, cosine distance, and scaled Euclidean distance. It is expected that different distance metrics will yield different results. 

\begin{figure*}[t]
\centering
\includegraphics[width=0.74\textwidth]{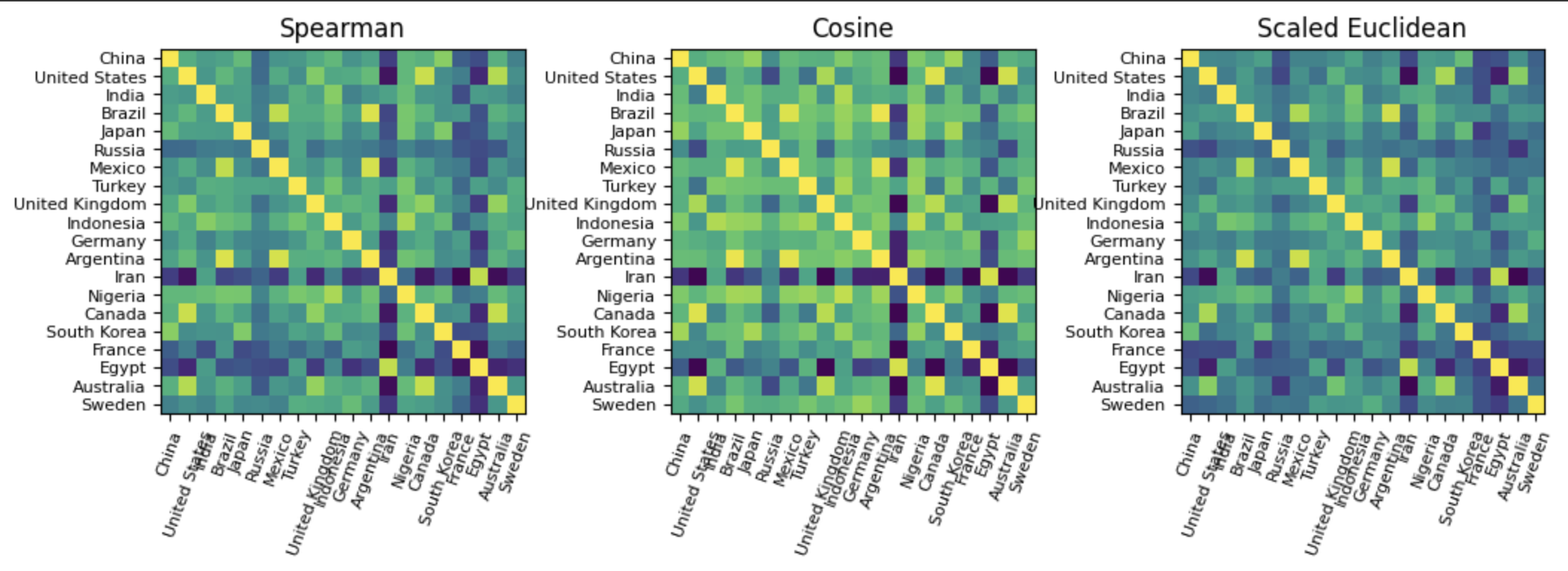}
\includegraphics[width=0.23\textwidth]{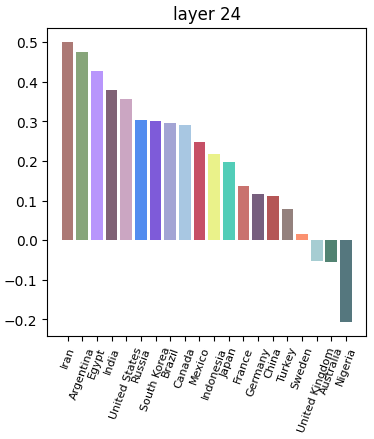}
\caption{\textbf{Left:} Distance between the GPT-Neo representation of countries, measured using Spearman correlation, cosine distance, and scaled Euclidean distance. \textbf{Right:} Kendall rank correlation $\tau$ for different countries in our dataset.}
\label{fig:gptneo_rsa}
\end{figure*}

To examine the alignment of the two representations, we used the Kendall rank correlation coefficient ($\tau$), which is calculated as: 
\begin{equation}
    \footnotesize
    \tau=\frac{(\#~{\mathrm{concordant~pairs}})-(\#~\mathrm{disconcordant~pairs})}{\mathrm{total}~\#~\mathrm{of~ pairs}}
\end{equation}
Specifically, we obtained one $\tau$ for each country. A higher $\tau$ suggests a higher alignment between the two spatial representations of a country. To check for statistical significance, we also computed the p-values for each country's $\tau$ value. To obtain the LLM activation representations' correlation to real geographical country representations, we took the mean of $\tau$ values for all countries.

\subsection{Results}
On the left of Figure \ref{fig:gptneo_rsa}, we show the heatmaps for the distance between the GPT-Neo representations of countries using three different distance metrics. It appears that countries that are closer to each other in real geographical locations are more correlated in their GPT-Neo hidden representations. For example, the US is close to Canada and far away from Iran in terms of longitudes and latitudes. Similar observations are made in the heatmaps, where Canada has the brightest yellow color and Iran has the darkest blue color for the row for the US. This demonstrates that GPT-Neo appears to be effective in capturing the geographical relationships among countries, with a high alignment of actual geographical positions of countries and spatial representations generated by GPT-Neo. 

In addition, we also present the Kendall rank correlation $\tau$ for different countries on the right of Figure \ref{fig:gptneo_rsa}. Most countries have positive $\tau$ values, indicating a good alignment between the two spatial representations for these two countries. The average $\tau$ values across all 100 countries are summarized in Table \ref{tab:tau_val}. In terms of p-values, 5 out of 20 countries' (i.e. 25\%) $\tau$ are statistically significant at 0.05 significance level. The RSA provides some explanations to the LLM representations of countries in terms of space, though these findings are not highly robust or conclusive. We explore probing in the next section, which offers stronger insights on learned LLM spatial representations.
\begin{table}[h]
    \centering
    \begin{tabular}{||c|c||}
    \hline\hline
    Similarity Metric    &  Average $\tau$\\\hline
    Spearman Correlation  & 0.202 \\\hline
    Cosine Distance & 0.207 \\ \hline
    Euclidean & 0.213
    \\\hline

    \end{tabular}
    \caption{Average $\tau$ coefficient of countries using different representation similarity metrics}
    \label{tab:tau_val}
\end{table}

\section{Linear and Non-linear Probing}
Another common approach for studying the learned representations in neural network is probing~\cite{alain2016understanding}, which trains an auxiliary classifier on the intermediate activation of a network to predict certain property of the input. A high prediction accuracy indicates high mutual information between the neural network's learned representations and the chosen property of its input~\cite{pimentel2020information}.

To study whether the LLM's learned representations encode the spatial property of its input, we trained probing regressors on the hidden states after each transformer layer of DeBERTa-v2 and GPT-Neo. The probing regressors map the intermediate hidden states of a city name to its real geographical coordinates. Formally, given $\mathbf{H}\in\mathbb{R}^{n\times d}$ is the intermediate hidden states of all cities from one transformer layer, and $\mathbf{Y}\in\mathbb{R}^{n\times 2}$ is the latitude and longitude of all cities, we searched for a function $f_\theta: \mathbb{R}^d\rightarrow\mathbb{R}^2$ s.t. the total residual is minimized. 
\begin{equation}
    \label{eqn: obj_fn}\theta=\argmin\limits_\theta\sum\limits_{i=0}^n\mathcal{L}\left(f_\theta(\mathbf{h}_i),\mathbf{y}_i\right)
\end{equation}
where $d$ is the dimension of the LLM's hidden states, $n$ is the total number of samples, and $\theta$ is the parameter of the probe. We repeated this process for each layer in DeBERTa-v2 and GPT-Neo.

\subsection{Loss Function}
In previous works, the Mean Squared Error (MSE) loss (Eqn. \ref{eqn: mse_obj_fn}) has been the common choice for the objective function of probing models \cite{abdou2021can,gurnee2023language}. 
\begin{equation}
\label{eqn: mse_obj_fn}
    \mathcal{L}_{\mathrm{MSE}}(\theta;\hat{\mathbf{Y}},\mathbf{Y})=\sum\limits_{i=0}^n|\hat{\mathbf{y}}_i-\mathbf{y}_i|^2=\sum\limits_{i=0}^n|f_\theta(\mathbf{h}_i)-\mathbf{y}_i|^2
\end{equation}However, we found a few drawbacks in MSE that make it suboptimal in this task working with latitude and longitude. One major problem is that MSE neglects the periodicity in longitude. For example, 180° east and 180° west are identical locations in terms of longitudinal coordinates, while they are viewed differently by MSE loss. Furthermore, MSE assumes Euclidean space, failing to consider the spherical shape of the earth's surface. For example, near the equator, two points with the same difference in longitude have a greater actual distance between them, whereas in high-latitudes, this difference is relatively smaller. Taking these into account, we proposed a novel loss function, which calculates the distance between two location on a sphere. We named it GeoDist. Formally, given two points $p_1(\varphi_1,\lambda_1),p_2(\varphi_2,\lambda_2)$ specified by latitude $\varphi_1,\varphi_2$ and longitude $\lambda_1,\lambda_2$, their distance apart is given by the Haversine equation
\begin{align}
    &\mathcal{H}(\theta) = \sin^{2}(\frac{\theta}{2}) \\
    &\mathcal{H}(p_1, p_2) = \mathcal{H}(\Delta \varphi) +  \cos \varphi_1 \cos \varphi_2 \mathcal{H}(\Delta \lambda) \nonumber \\ 
    & \mathrm{GeoDist}(p_1,p_2)=2r\cdot \arcsin (\sqrt{\mathcal{H}(p_1, p_2)}) \nonumber 
\end{align}
\cite{Gade2010NonSR}, where $r$ is the radius of the sphere. The proposed GeoDist loss is then 
\begin{equation}
\label{eqn: geodist_loss}
    \mathcal{L}_{\mathrm{GeoDist}}(\theta;\hat{\mathbf{Y}},\mathbf{Y})=\sum\limits_{i=0}^n\mathrm{GeoDist}
(f_\theta(\mathbf{h}_i)-\mathbf{y}_i)
\end{equation}
In our work, we trained probing models using both MSE and GeoDist loss, and compared the performance of models in terms of regression accuracy (Section \ref{subsec: lin_probe}) and significance in downstream tasks (Section \ref{sec: causality}). 

\subsection{Linear Representations}
\label{subsec: lin_probe}
We first restricted $f$ to only \emph{linear functions}, $f\in\mathcal{L}(\mathbb{R}^d,\mathbb{R}^2)$. This is equivalent to find a linear transformation kernel $T\in\mathbb{R}^{d\times2}$ that maps activation to latitude and longitude. For MSE loss, the maximum likelihood estimator (MLE) of $T$ has a closed form  
\begin{equation}
    T=(\mathbf{H}^T\mathbf{H})^{-1}\mathbf{H}^T\mathbf{Y}
\end{equation}
For GeoDist loss, the model parameter is optimized via stochastic gradient descent (SGD), updated iteratively as $\theta=\theta-\alpha\nabla \mathcal{L}_{\mathrm{GeoDist}}(\theta;\mathbf{H},\mathbf{Y})$ 
until convergence.

\begin{figure}[h]
\centering
\includegraphics[width=0.22\textwidth]{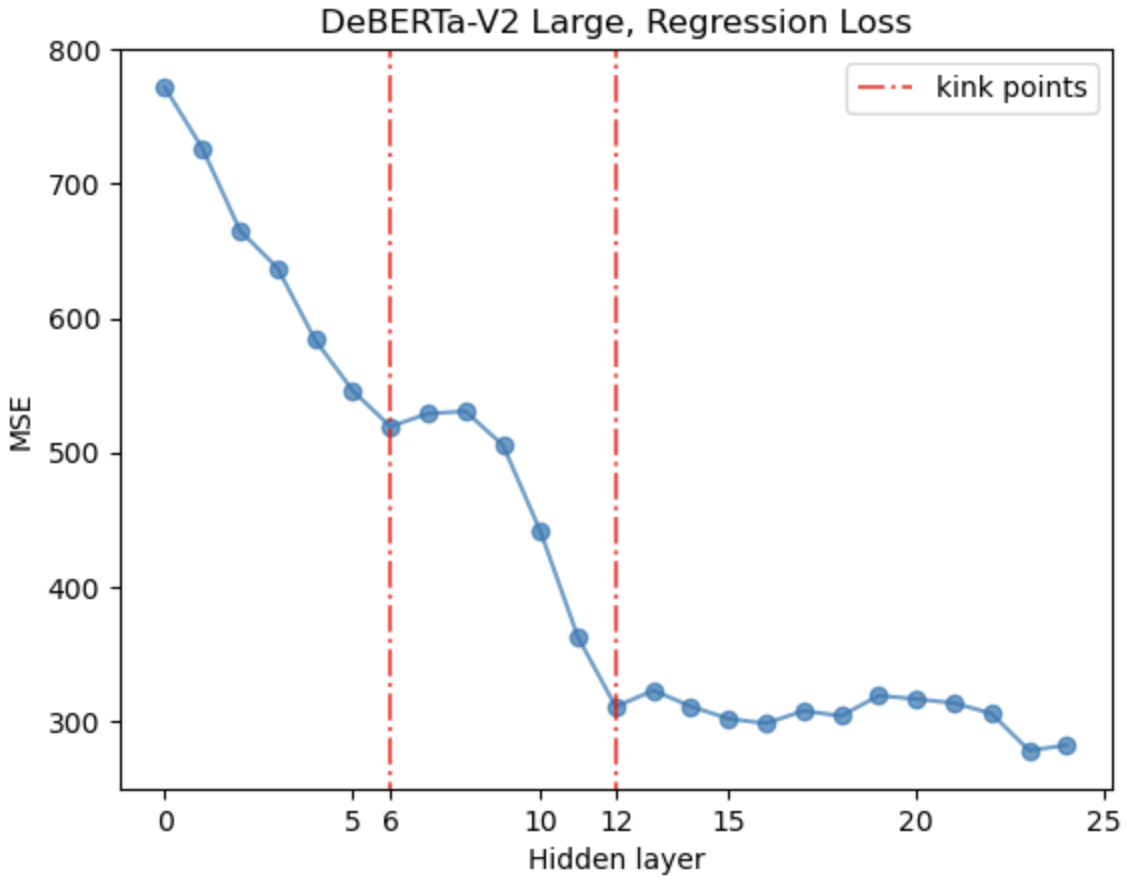}
\includegraphics[width=0.22\textwidth]{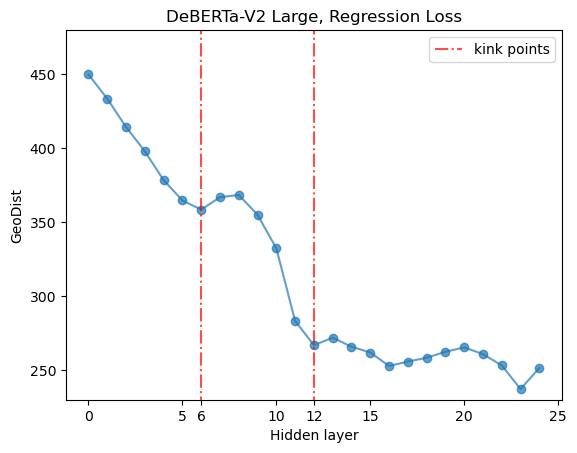}
\caption{Linear probe performance on DeBERTa-v2 across transformer layers\\
\textbf{Left:} MSE loss \textbf{Right:} GeoDist loss}
\label{fig:deberta_mse}
\end{figure}

For each set of activation, the mean validation loss from 10-fold cross-validation was computed to evaluate the predictive ability of the activation~\cite{crossval1974,McLachlan2004AnalyzingMG}. A low out-of-sample error signals a consistent, learnable pattern that decodes geographical coordinates from activation. Such coherence in these two representation spaces is possibly the result of a spatial world model that LLMs developed internally. See Figure \ref{fig:deberta_mse} for the average loss of linear probe fitted on activation of transformer block 1 to 24 in DeBERTa-v2.
\\
\\
\begin{figure*}[t]
\centering
\includegraphics[width=1\textwidth]{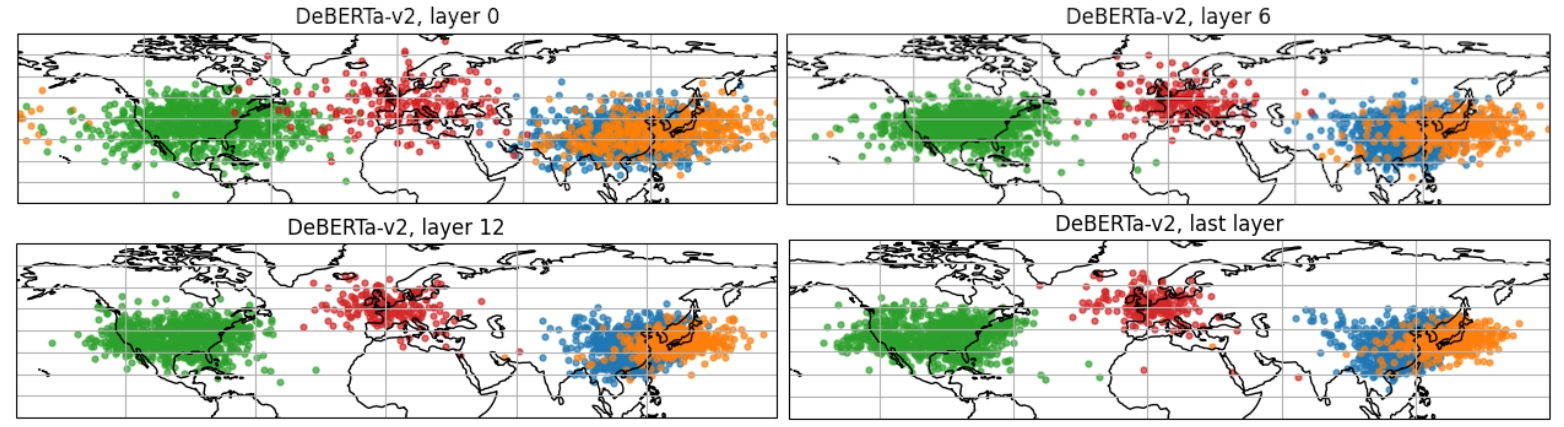}
\caption{DeBERTa-v2 Mapping v.s. Layer Depth. Each colored dot represents a city. Its longitude and latitude are predicted by a linear model that takes DeBERTa-v2's internal representations of the city names as input. \\ \centerline{(Blue: China; Orange: Japan; Green: US; Red: UK)}}
\label{fig:deberta_maps}
\end{figure*}
\begin{figure*}[t]
\centering
\includegraphics[width=1\textwidth]{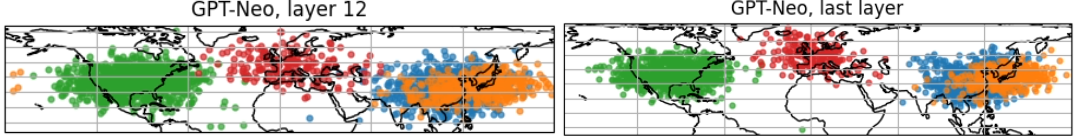}
\caption{GPT-Neo Mapping v.s. Layer Depth}
\label{fig:gptneo_maps}
\end{figure*}
From Figure~\ref{fig:deberta_mse}, we observed similar results for both loss functions. Specifically, the regression loss obtained by the linear probe is lower in later transformer layers compared to earlier ones. It suggested that the features or representations learned in the later transformer layers contain more relevant space information. In practice, we noted that the performance of the linear probe stabilized around the 12th hidden layer. This suggests that, beyond this layer, there was no significant contribution in capturing further relevant information. 
\begin{figure*}[h!]
\centering
\includegraphics[width=0.5\textwidth]{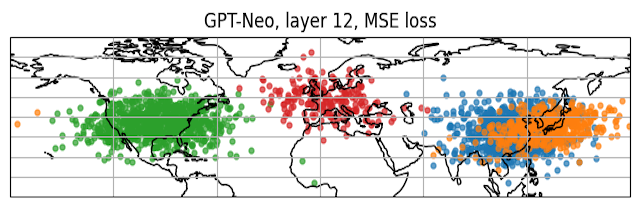}
\includegraphics[width=0.49\textwidth]{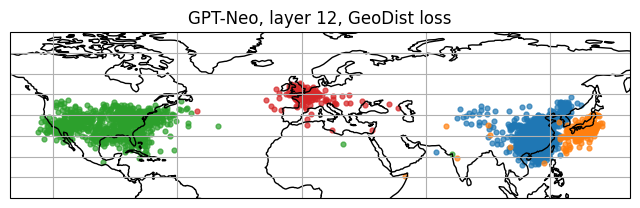}
\caption{GPT-Neo Probing v.s. Loss Function}
\label{fig:gptneo_maps_loss}
\end{figure*}
In Figure~\ref{fig:deberta_maps}, we plotted the predictions of the DeBERTa-v2 linear probe for longitude and latitude of major cities in the US, the UK, China and Japan on a world map. We may conclude that the 12th and last transformer layers have better predictions than the 0th and the 6th. While China and Japan were closely positioned together, the overall geographical representations are deemed satisfactory. Similar trends have been observed for GPT-Neo, as shown in Figure~\ref{fig:gptneo_maps}, where the predictions for longitude and latitude exhibit notable similarities between layer 12 and the last layer.

The probing results are consistent across different choices of loss functions. Probing models trained on GeoDist loss also achieved low out-of-sample error, signaling that the mutual information shared between LLM activation and real-world geography goes deeper than the value of latitude and longitude, which is no more than a 2-dimensional numerical representations of geographical locations, and is rooted in the true spatial world model. Figure \ref{fig:gptneo_maps_loss}
shows the visualizations of probing results of models trained on both losses. For the same model architecture and activation used, the model trained with GeoDist loss is able to reconstruct the world map that aligns better with the real world humans perceive. In Figure \ref{fig:gptneo_maps_loss} right, all countries have their cities concentrated in the region of their real territory, and are clearly separable from other countries. It is even possible to vaguely see the border of the U.S. and China on the map. Whereas all countries are just unshaped blobs of dots; plus China and Japan intermingle with each other on the map on the left of Figure \ref{fig:gptneo_maps_loss}. This could be a even more conclusive evidence that the language model spontaneously learns the spatial world model since its activation encode not only the values of latitude and longitude, but the geometry of the world as well. 

\subsection{Non-linear Probe}
Unlike previous works~\cite{abdou2021can,gurnee2023language} that restricted probes to be linear models only, we took one more step by considering more complex probes as well. The objective function is still Equation (\ref{eqn: obj_fn}), but $f_\theta\in\mathbb{R}^d\rightarrow\mathbb{R}^2$ can be arbitrary functions now. 

One class of simple non-linear models is multi-layer feed-forward neural networks (FFNN). Introducing additional layers allows the model to learn hierarchical representations and to compose lower-level features into higher-level abstractions~\cite{Bengio2013feature}. Compared to linear models, deep FFNN is able to better recognize the complex non-linear representations in the training data with less parameters~\cite{Bengio2007LearningDA}, thus exponentially more accurate than linear models in certain tasks~\cite{Rolnick2017ThePO}. 

We trained probes of $l$ layers and $k$ dimensions per layer on both DeBERTa and GPT-Neo activation. Different combinations of hyper-parameter $l$ and $k$ are explored via grid search. 
\begin{figure}
\begin{minipage}[b]{.45\linewidth}
\includegraphics[width=\linewidth]{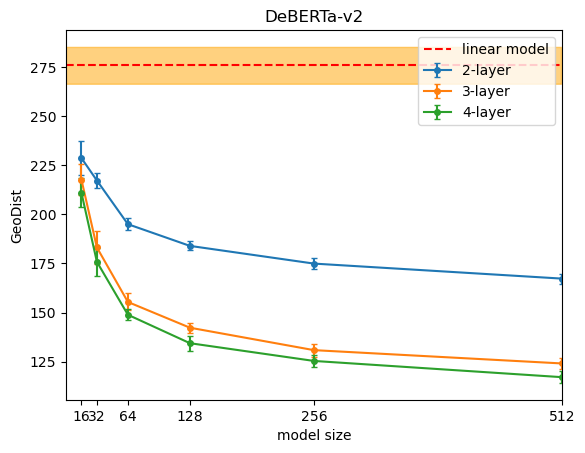}
\end{minipage}\hfill
\begin{minipage}[b]{.45\linewidth}
\includegraphics[width=\linewidth]{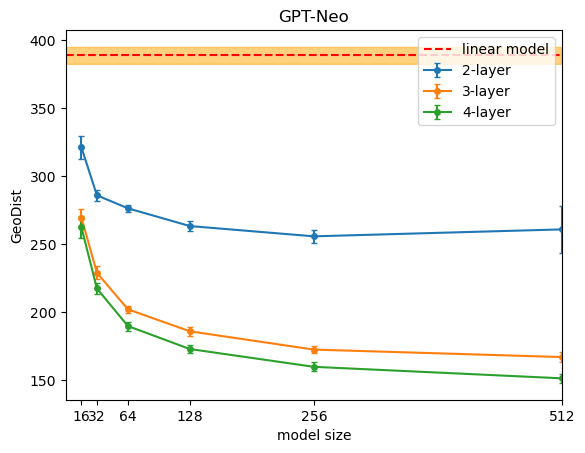}
\end{minipage}
\caption{Validation performance (GeoDist loss) of non-linear probe with different hidden neurons trained on the activation from the 12th transformer block of both DeBERTa-v2 and GPT-Neo model}
\label{fig:mse_vs_modelsize}
\end{figure}
Figure \ref{fig:mse_vs_modelsize} plots the impact of model size on the performance when probing the activation from the 12th hidden layers in both DeBERTa and GPT-Neo. The same experiment was repeated on the activation from every layer, and similar results were observed. For both models, we found that more complex probes consistently outperformed linear probes, and the boost in performance is more significant for GPT-Neo activation. For the same model depth, the validation loss drops drastically at first as the number of neurons per hidden layer increases but plateaus quickly. For the same hidden dimension, deeper probes exhibit stronger predictive ability for $l<3$, after which the marginal gain of stacking more layers becomes negligible.   

In general, we observed a discernible ceiling on the predictive power of the probes. But for both DeBERTa and GPT-Neo activation, linear probes are far from saturated, which differs from conclusions in previous work that non-linear probe do not outperform linear one~\cite{gurnee2023language}. 

One major criticism on non-linear probes is that overly expressive probes may overfit on the activation dataset - although the spatial information is decodable from the activation, it's unclear if such spatial world model is indeed used by the LLM~\cite{Ravichander2020probing}. We address such concerns about the real causality in Section \ref{sec: causality}.

\section{Causality of Spatial Representations}
\label{sec: causality}
\begin{figure*}[t]
\centering
\includegraphics[width=1\textwidth]{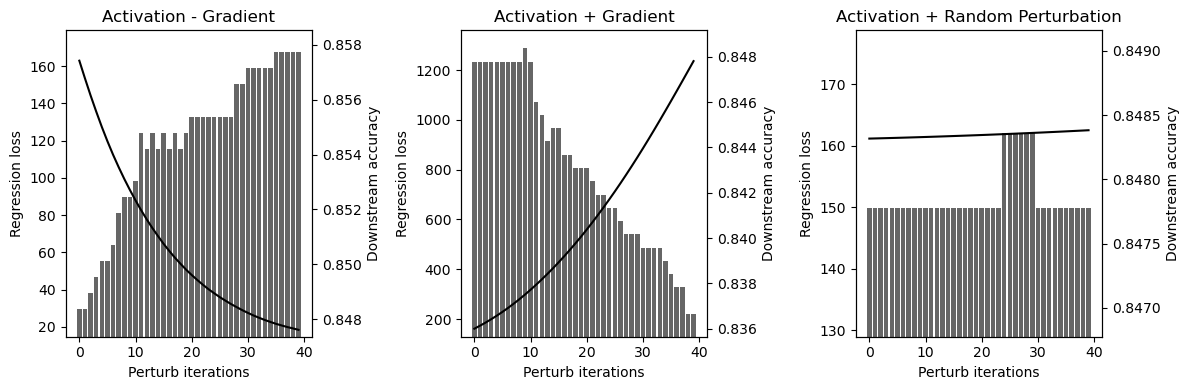}
\caption{GPT-Neo regression loss and downstream classfication accuracy v.s. number of iterations.}
\label{fig:gpt_reg_acc}
\end{figure*}
\begin{figure*}[t]
\centering
\includegraphics[width=1\textwidth]{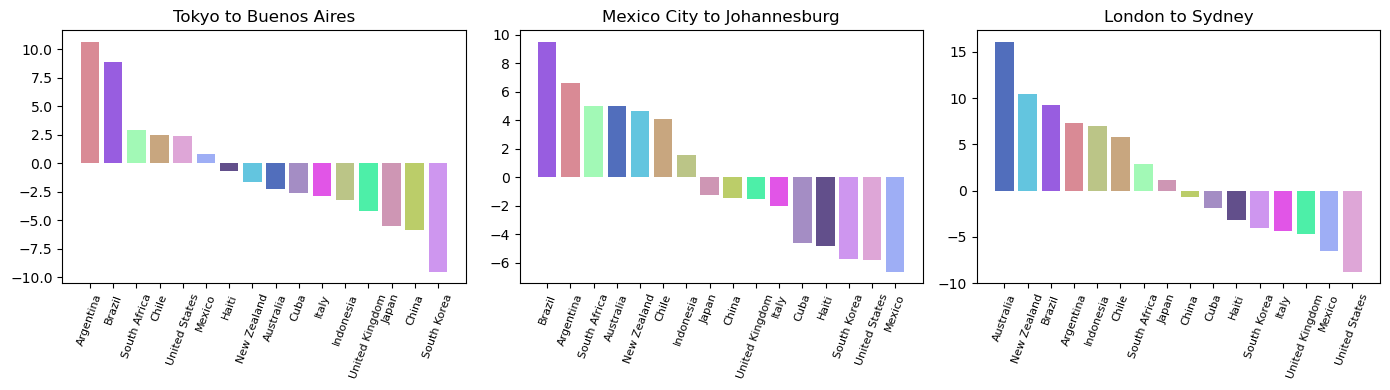}
\caption{GPT-Neo classification logit change under targeted perturbation}
\label{fig:gpt_targeted_pert}
\end{figure*}

To study the causal effect of spatial representations on the model's output, we modified the intermediate hidden states~\cite{giulianelli2018under, tucker2021if} of location words and measured changes in the model's behavior and downstream task performance. 

\subsection{Causal Role in a Spatial Downstream Task}\label{country_pred}
We started with a simple geospatial task --- we trained a classifier on the last layer hidden states to predict the country of a given city, and we measured how the classifier prediction changed as we modified the intermediate hidden states of the city using the gradients from probing classifiers. 

If the latitude and longitude representations are used in country prediction, modifying them with respect to the true city longitudes and latitudes would lead to an improvement in downstream classification. As a baseline comparison, we conducted random perturbation and gradient-opposite perturbation. In a more extreme scenario, if we perturb the transformer layer by back-propagating a different city's location, we would anticipate a downstream prediction corresponding to the country of the perturbed city. For instance, if the true label of the transformer layer for the linear probe is the city Tokyo, and we perturb the transformer layer using the longitudes and latitudes of Buenos Aires, we expect the downstream prediction to shift to Argentina instead of Japan.


\subsubsection{Results}\label{country_pred_results}
As shown in the first plot in Figure \ref{fig:gpt_reg_acc}, we optimized the intermediate hidden states using the gradients of the trained linear probe $f_{\theta}$, so the longitude and latitude representations inside GPT-Neo more precisely match with the true longitude and latitude of the city, $\mathbf{h}_{t+1} = \mathbf{h}_{t} - \frac{\partial \mathcal{L}(f_{\theta}(\mathbf{h}_t), \mathbf{Y})}{\partial \mathbf{h}_{t}}$ where $h$ is the intermediate hidden state, $\mathbf{Y}$ is the true longitude and latitude, and $\mathcal{L}(\cdot)$ is either Eqn \ref{eqn: mse_obj_fn} or \ref{eqn: geodist_loss} depending on the choice of loss function. As the perturbation amplitude increases, the hidden representations of each city names keeps being mapped closer to its true geographical location, and the accuracy of the downstream country classification task increases monotonically at the same time. In contrast, when perturbing the intermediate hidden states by adding gradient or introducing random noises, we observed a decrease in downstream task accuracy and no change, respectively. There exists a clear consistent and robust negative correlation between the probing loss and the downstream classification accuracy, supporting the utility of longitude and latitude representations for predicting their respective countries. 
 
Furthermore, we performed targeted perturbation on selected cities. For example, in the first plot of Figure \ref{fig:gpt_targeted_pert}, we perturbed the hidden states of Tokyo by aligning it with the longitude and latitude of Buenos Aires in the linear probe. Subsequently, upon feeding the perturbed hidden states into the downstream classifier, sensible shifts in logits were observed -- the logits for Southern American countries, such as Argentina and Brazil, increased, while those for Asian countries, including Japan, China, and South Korea, decreased, although the logit change was not large enough to alter the final prediction. This further confirmed the utilization of longitude and latitude information in country predictions.
 
\begin{figure}[h]
\centering
\includegraphics[width=0.45\textwidth]{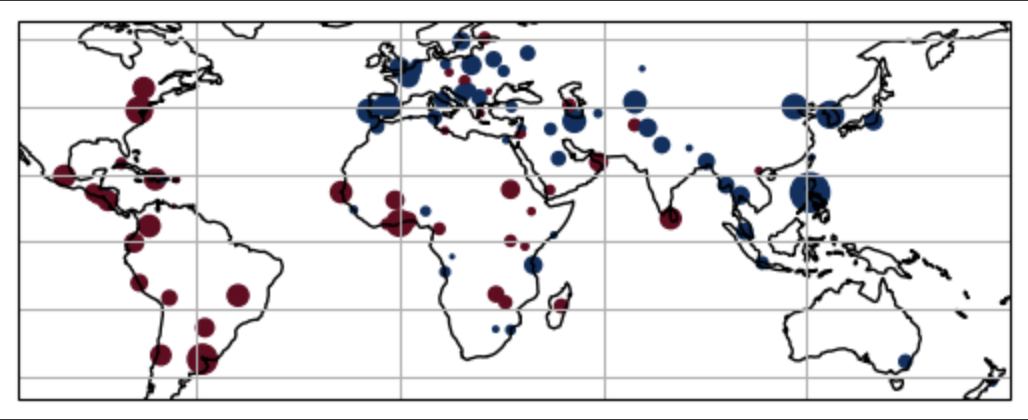}
\caption{GPT-Neo classification logit change under targeted perturbation (world map view)}
\label{fig:gpt_logit_change}
\end{figure}

Similar observation is made when plotting the change in logits in the world map view for targeted perturbation. Taking the example of perturbing the hidden states of Tokyo with Buenos Aires in Figure \ref{fig:gpt_logit_change}, we observed significant decrease in the logits on the right of the world map, including Asian countries (shown as blue circles) and significant increase in the logits on the left of the world map, including South American countries (shown as red circles). 
\\ This shows a shift in country prediction from Asian countries to South American countries after applying perturbations to city hidden states, providing evidence that LLMs learn the world model of space internally rather than simply memorizing the training data. 

\subsubsection{Statistical Significance Test}
To verify that the change in classification accuracy shown in Figure \ref{fig:gpt_reg_acc} is not the result of random fluctuation, we computed the change in true-label-class logit before and after intervention, $\Delta_{\mathrm{logit}}^{(i)}=\delta_{y_ij} (\mathbf{f}_{cls}(\mathbf{\Tilde{h}}_i)-\mathbf{f}_{cls}(\mathbf{h}_i))^j$, where $\Tilde{\mathbf{h}}$ is the activation after intervention and $\mathbf{f}_{cls}$ is the classification model. The mean of true-label-class logit change of all samples is expected to be larger than 0 if the change in classification accuracy is indeed the result of intervention, meaning that the geospatial knowledge decoded by probes is utilized by LLM, and 0 otherwise. \emph{Z-test} was performed for the mean logit change to verify its statistical significance. 
\begin{figure}[h]
    \centering
    \includegraphics[width=0.4\textwidth]{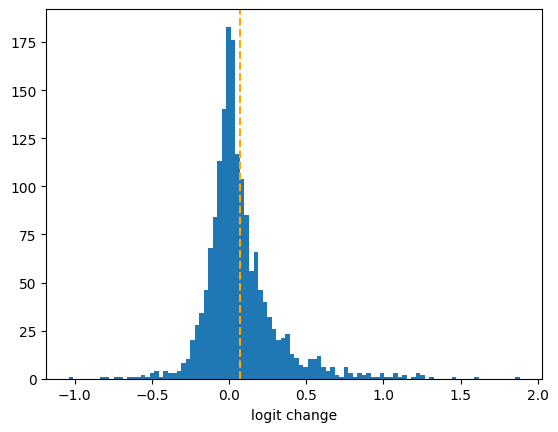}
    \caption{Histogram of true-label-class logit change after intervention\\
    mean: 0.122, $p$-value: 4.43e-33}
    \label{fig:enter-label}
\end{figure}
\\
At 0.05 significance level, we may conclude that the change in logit is statistically significant.

\subsubsection{Compare GeoDist Loss to MSE}
The above intervention experiment was repeated on probes trained by both MSE and GeoDist loss. We were also interested in seeing if switching to the novel loss function would impact the downstream task differently. 
\begin{figure}
    \centering
    \includegraphics[width=0.23\textwidth]{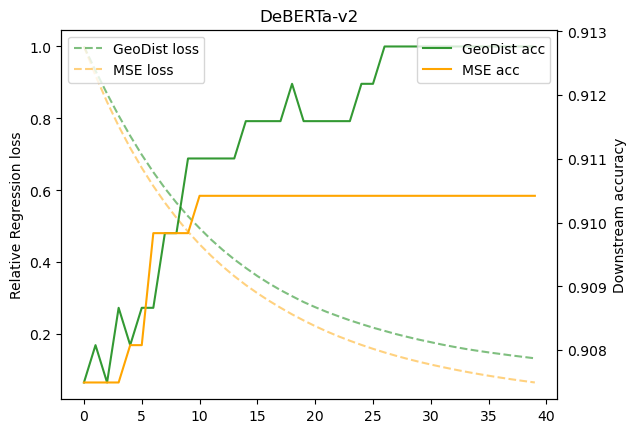}
    \includegraphics[width=0.22\textwidth]{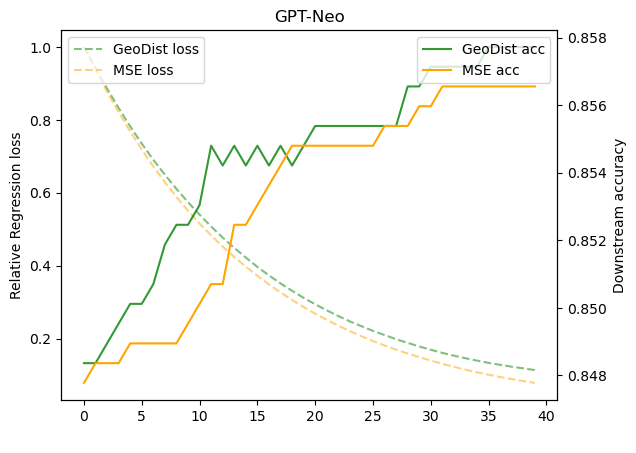}
    \caption{Correlation between probing loss and downstream accuracy for both loss functions \\ 
    Probe: FFNN $l=3,k=128$ \\
    Activaiton: \textbf{Left:} DeBERTa $12^{th}$ layer; \textbf{Right:} GPT-Neo $12^{th}$ layer}
    \label{fig:cmp_mse_geodist}
\end{figure}
As shown in Figure \ref{fig:cmp_mse_geodist}, the negative correlation between probing loss and classification accuracy is consistent for both loss functions. It can also be observed that the BERT-based classifier gains an additional 0.025 accuracy if the activation is aligned with real geography by GeoDist, compared to MSE. For GPT-based classifier, both loss functions offer similar level of boost in downstream accuracy, but the curve of GeoDist model is steeper at small intervention amplitude, meaning that LLM is more sensitive to GeoDist - a small improvement in the alignment between LLM activation and real geospatial knowledge, measured by GeoDist is sufficient to enhance downstream performance by a notable amount. 

For both DeBERTa and GPT-Neo, we observed that perturbing the activation to make it closer to true location in real-world distance is more effective in improving the LLM downstream performance, compared to simply aligning it to the value of latitude and longitude. This observation suggests that LLM is aware of not only the numerical attribute (latitude, longitude) of locations, but also knowledge about the real geometry of earth, which further consolidates our finding that LLM internally developed real-world representations of space, going beyond statistical correlation. 

\subsection{Causal Role in Next Word Prediction}
With positive results regarding causality in the country prediction task, we came to the original question we wanted to answer - are the spatial representations actually used by LLMs to predict the next word? 

We employed the following prompt for each city in the dataset: 
\begin{itemize}
   \item[] \texttt{<city admin\_name> is located in the country of}
\end{itemize}
The ground-truth label is the country associated with that city. If the label consists of multiple tokens, we consider the first token as the label. To illustrate, if the city is \texttt{Paris}, we would expect the next word to be \texttt{France} with a high probability. 

\begin{figure}[h]
\centering
\includegraphics[width=0.45\textwidth]{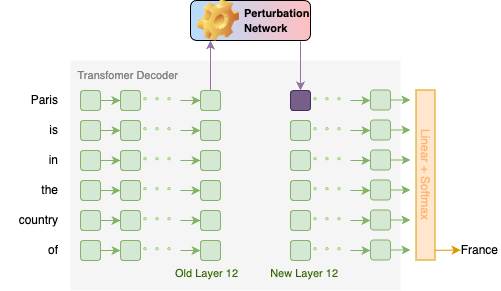}
\caption{Perturbing the next word prediction of GPT-Neo. The activation of the last token of the city name from 12th layer was extracted and perturbed with the gradients of the probing network.}
\label{fig:next_word_perturb}
\end{figure}

To perform the causal intervention, we extracted the activation of the last token of the city name from 12th layer of GPT-Neo, perturbed it with the same technique as discussed in Section \ref{country_pred}. We then substituted the original activation with the perturbed counterpart in 12th layer and proceeded to pass through the subsequent layers and got the next word prediction result on the last token. This process is depicted in Figure \ref{fig:next_word_perturb}.
\\
\\
To assess the impact of this causal intervention on next word prediction, we measured the following three metrics on the test set:
\begin{enumerate}
    \item Change in accuracy
    \item Change in top-5 accuracy, where top-5 accuracy is the proportion of times the ground-truth label appears in the top-5 predictions
    \item Change in the logit of the ground-truth label
\end{enumerate}


   
\subsubsection{Results}
\begin{figure}[h]
\centering
\includegraphics[width=0.5\textwidth]{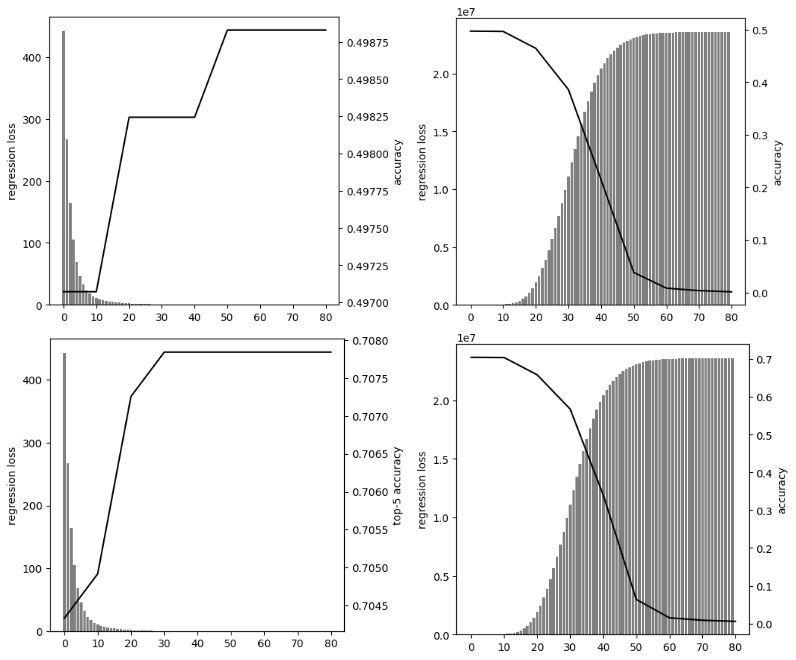}
\caption{Metric Change v.s. Number of Iterations, \\ \textbf{Top Left:} Accuracy, Gradient Descent, $\delta=0.0018$; \\ \textbf{Top Right:} Accuracy, Gradient Ascent, $\delta=0.0035$; \\ \textbf{Bottom Left:} Top-5 Accuracy, Gradient Descent, $\delta=-0.4959$; \\ \textbf{Bottom Left:} Top-5 Accuracy, Gradient Ascent, $\delta=-0.6984$}
\label{fig:nextword_acc}
\end{figure}

\paragraph{Change in accuracy and top-5 accuracy\\}
We performed gradient descent and ascent respectively using the gradients from the probing network on the activation for 80 iterations. The change in accuracy, top-5 accuracy and regression loss with the number of iterations are shown in Figure \ref{fig:nextword_acc}. The negative correlation between regression loss of the probe with both accuracy and top-5 accuracy is evident, similarly as what we observed in Section \ref{country_pred_results}. 

However, it's noteworthy that the reduction in accuracy during gradient ascent had a substantially larger magnitude than the increase observed in gradient descent. In gradient descent, there was only a tiny improvement in accuracy and top-5 accuracy of 0.0018 and 0.0035 respectively. However, in gradient ascent, both metrics declined to nearly zero. Our conjecture is that the LLM was already close to its optimal capacity, so the augmented spatial information in the activation after gradient descent didn't help much with the prediction. However, removing this information via gradient ascent led to a drastic degradation in LLM performance, supporting the utilization of spatial information in LLM prediction and establishing the causal link.

\begin{figure}[h]
\centering
\includegraphics[width=0.5\textwidth]{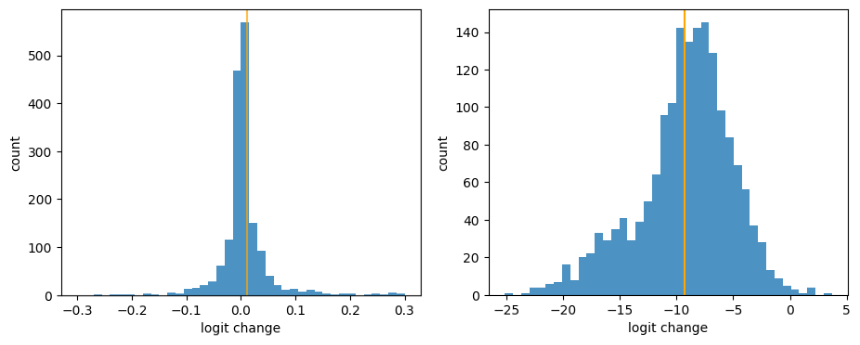}
\caption{Histogram of Logit Change, \\ \textbf{Left:} Gradient Descent, mean=0.0118, p-value=0; \\ \textbf{Right:} Gradient Ascent, mean=-9.25, p-value=0}
\label{fig:logit_change}
\end{figure}

\paragraph{Change in the logit of the ground-truth label\\}
To further verify the causal link, we examined the logit change of the ground-truth label after perturbation. Histograms depicting the logit change for both gradient descent and gradient ascent are presented in Figure \ref{fig:logit_change}. There is an average increase of 0.0118 for gradient descent, and an average decrease of 9.25 for gradient ascent, both highly significant with a p-value close to zero. The magnitude of the change observed in gradient ascent is much larger than that in gradient descent, for the same reason as discussed in the preceding section.

\begin{figure}[h]
\centering
\includegraphics[width=0.5\textwidth]{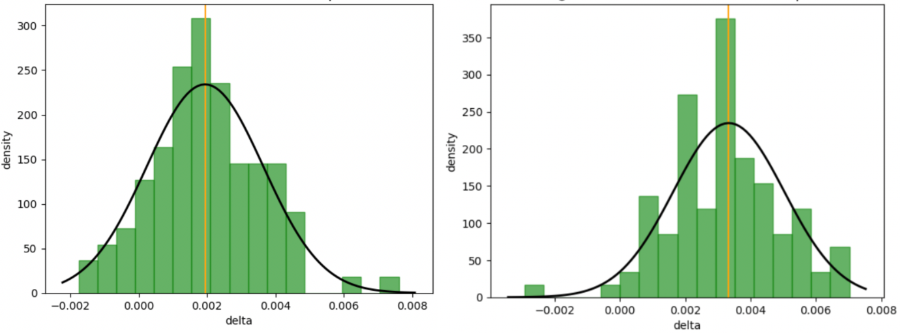}
\caption{\textbf{Left:} Histogram of Change in Accuracy, mean=0.0019, p-value=0; \textbf{Right:} Histogram of Change in Top-5 Accuracy, mean=0.0033, p-value=0}
\label{fig:sig_test}
\end{figure}

\paragraph{Statistical significance of changes\\} 
Given that the improvements in accuracy and top-5 accuracy are very small, we conducted a \emph{Z-test} to assess the significance of these improvements. In each experiment, we performed gradient descent on the activation for 80 iterations and recorded the final changes in accuracy and top-5 accuracy. This experiment was repeated for 100 times with different seeds. The distributions of the changes are shown in Figure \ref{fig:sig_test}. Despite the small magnitude, the improvements in accuracy and top-5 accuracy are both significant, with a p-value close to zero, further reinforcing our conclusion regarding the causal impact.

\section{Conclusion}
Through representational similarity analysis and probing, we have shown that both DeBERTa-v2 and GPT-Neo learned the spatial representations of a set of location words through pretraining. Intervention on the country prediction task further demonstrated that the spatial representations have causal effects on the model's output. By aligning the learned spatial representations with the true geographical location of a city, we were able to improve the accuracy of the country prediction from GPT-Neo's learned representations. Our results provided additional insights why LLMs can be easily fine-tuned for solving geospatial tasks or answering geospace-related questions in a zero-shot setting.
\label{conclusion}

\section{Future Work}
\label{future_work}
We have shown that the internal representations of LLMs encode spatial property of the location words. Intervention experiments on the activation suggests that the spatial representations also have causal influence on the model's downstream task performance. However, a mystery is yet to be solved. While certain words, such as city names, have inherent geographical property, many words contain no geographical information. ``Human", for example, is present almost anywhere on the earth and cannot be mapped to a specific location naturally. Where would these words be on our identified longitude and latitude dimensions? What effect would perturbation on spatial representations bring to the downstream task if the task doesn't rely on geographical knowledge? As part of the future work, we will conduct additional experiments to answer the questions above.

\bibliographystyle{acl_natbib}
\bibliography{anthology,acl2021}

\begin{thebibliography}{35}
\expandafter\ifx\csname natexlab\endcsname\relax\def\natexlab#1{#1}\fi

\bibitem[{Abdou et~al.(2021)Abdou, Kulmizev, Hershcovich, Frank, Pavlick, and S{\o}gaard}]{abdou2021can}
Mostafa Abdou, Artur Kulmizev, Daniel Hershcovich, Stella Frank, Ellie Pavlick, and Anders S{\o}gaard. 2021.
\newblock Can language models encode perceptual structure without grounding? a case study in color.
\newblock \emph{arXiv preprint arXiv:2109.06129}.

\bibitem[{Alain and Bengio(2016)}]{alain2016understanding}
Guillaume Alain and Yoshua Bengio. 2016.
\newblock Understanding intermediate layers using linear classifier probes.
\newblock \emph{arXiv preprint arXiv:1610.01644}.

\bibitem[{Bender and Koller(2020)}]{bender2020climbing}
Emily~M Bender and Alexander Koller. 2020.
\newblock Climbing towards nlu: On meaning, form, and understanding in the age of data.
\newblock In \emph{Proceedings of the 58th annual meeting of the association for computational linguistics}, pages 5185--5198.

\bibitem[{Bengio(2007)}]{Bengio2007LearningDA}
Yoshua Bengio. 2007.
\newblock \href {https://api.semanticscholar.org/CorpusID:207178999} {Learning deep architectures for ai}.
\newblock \emph{Found. Trends Mach. Learn.}, 2:1--127.

\bibitem[{Bengio et~al.(2012)Bengio, Courville, and Vincent}]{Bengio2013feature}
Yoshua Bengio, Aaron~C. Courville, and Pascal Vincent. 2012.
\newblock \href {http://arxiv.org/abs/1206.5538} {Unsupervised feature learning and deep learning: {A} review and new perspectives}.
\newblock \emph{CoRR}, abs/1206.5538.

\bibitem[{Bisk et~al.(2020)Bisk, Holtzman, Thomason, Andreas, Bengio, Chai, Lapata, Lazaridou, May, Nisnevich et~al.}]{bisk2020experience}
Yonatan Bisk, Ari Holtzman, Jesse Thomason, Jacob Andreas, Yoshua Bengio, Joyce Chai, Mirella Lapata, Angeliki Lazaridou, Jonathan May, Aleksandr Nisnevich, et~al. 2020.
\newblock Experience grounds language.
\newblock \emph{arXiv preprint arXiv:2004.10151}.

\bibitem[{Black et~al.(2021)Black, Gao, Wang, Leahy, and Biderman}]{gpt-neo}
Sid Black, Leo Gao, Phil Wang, Connor Leahy, and Stella Biderman. 2021.
\newblock \href {https://doi.org/10.5281/zenodo.5297715} {{GPT-Neo: Large Scale Autoregressive Language Modeling with Mesh-Tensorflow}}.

\bibitem[{Brown et~al.(2020)Brown, Mann, Ryder, Subbiah, Kaplan, Dhariwal, Neelakantan, Shyam, Sastry, Askell et~al.}]{brown2020language}
Tom Brown, Benjamin Mann, Nick Ryder, Melanie Subbiah, Jared~D Kaplan, Prafulla Dhariwal, Arvind Neelakantan, Pranav Shyam, Girish Sastry, Amanda Askell, et~al. 2020.
\newblock Language models are few-shot learners.
\newblock \emph{Advances in neural information processing systems}, 33:1877--1901.

\bibitem[{Dan et~al.(2020)Dan, He, and Roth}]{dan2020understanding}
Soham Dan, Hangfeng He, and Dan Roth. 2020.
\newblock Understanding spatial relations through multiple modalities.
\newblock \emph{arXiv preprint arXiv:2007.09551}.

\bibitem[{Devlin et~al.(2018)Devlin, Chang, Lee, and Toutanova}]{devlin2018bert}
Jacob Devlin, Ming-Wei Chang, Kenton Lee, and Kristina Toutanova. 2018.
\newblock Bert: Pre-training of deep bidirectional transformers for language understanding.
\newblock \emph{arXiv preprint arXiv:1810.04805}.

\bibitem[{Elazar et~al.(2021)Elazar, Ravfogel, Jacovi, and Goldberg}]{elazar2021amnesic}
Yanai Elazar, Shauli Ravfogel, Alon Jacovi, and Yoav Goldberg. 2021.
\newblock Amnesic probing: Behavioral explanation with amnesic counterfactuals.
\newblock \emph{Transactions of the Association for Computational Linguistics}, 9:160--175.

\bibitem[{Feder et~al.(2021)Feder, Oved, Shalit, and Reichart}]{feder2021causalm}
Amir Feder, Nadav Oved, Uri Shalit, and Roi Reichart. 2021.
\newblock Causalm: Causal model explanation through counterfactual language models.
\newblock \emph{Computational Linguistics}, 47(2):333--386.

\bibitem[{{Gade}(2010)}]{Gade2010NonSR}
Kenneth {Gade}. 2010.
\newblock \href {https://doi.org/10.1017/S0373463309990415} {{A Non-singular Horizontal Position Representation}}.
\newblock \emph{Journal of Navigation}, 63(3):395--417.

\bibitem[{Giulianelli et~al.(2018)Giulianelli, Harding, Mohnert, Hupkes, and Zuidema}]{giulianelli2018under}
Mario Giulianelli, Jacqueline Harding, Florian Mohnert, Dieuwke Hupkes, and Willem Zuidema. 2018.
\newblock Under the hood: Using diagnostic classifiers to investigate and improve how language models track agreement information.
\newblock \emph{arXiv preprint arXiv:1808.08079}.

\bibitem[{Gurnee and Tegmark(2023)}]{gurnee2023language}
Wes Gurnee and Max Tegmark. 2023.
\newblock Language models represent space and time.
\newblock \emph{arXiv preprint arXiv:2310.02207}.

\bibitem[{He et~al.(2020)He, Liu, Gao, and Chen}]{he2020deberta}
Pengcheng He, Xiaodong Liu, Jianfeng Gao, and Weizhu Chen. 2020.
\newblock Deberta: Decoding-enhanced bert with disentangled attention.
\newblock \emph{arXiv preprint arXiv:2006.03654}.

\bibitem[{Hernandez et~al.(2023)Hernandez, Sharma, Haklay, Meng, Wattenberg, Andreas, Belinkov, and Bau}]{hernandez2023linearity}
Evan Hernandez, Arnab~Sen Sharma, Tal Haklay, Kevin Meng, Martin Wattenberg, Jacob Andreas, Yonatan Belinkov, and David Bau. 2023.
\newblock Linearity of relation decoding in transformer language models.
\newblock \emph{arXiv preprint arXiv:2308.09124}.

\bibitem[{Karney and Deakin(2010)}]{Karney_2010}
C.F.F. Karney and R.E. Deakin. 2010.
\newblock \href {https://doi.org/10.1002/asna.201011352} {F.w. bessel (1825): The calculation of longitude and latitude from geodesic measurements}.
\newblock \emph{Astronomische Nachrichten}, 331(8):852–861.

\bibitem[{Kriegeskorte et~al.(2008)Kriegeskorte, Mur, and Bandettini}]{Kriegeskorte2008}
Nikolaus Kriegeskorte, Marieke Mur, and Peter Bandettini. 2008.
\newblock \href {https://doi.org/10.3389/neuro.06.004.2008} {Representational similarity analysis - connecting the branches of systems neuroscience}.
\newblock \emph{Frontiers in Systems Neuroscience}, 2.

\bibitem[{LeCun(2022)}]{lecun2022path}
Yann LeCun. 2022.
\newblock A path towards autonomous machine intelligence version 0.9. 2, 2022-06-27.
\newblock \emph{Open Review}, 62.

\bibitem[{Li et~al.(2021)Li, Nye, and Andreas}]{li2021implicit}
Belinda~Z Li, Maxwell Nye, and Jacob Andreas. 2021.
\newblock Implicit representations of meaning in neural language models.
\newblock \emph{arXiv preprint arXiv:2106.00737}.

\bibitem[{McLachlan et~al.(2004)McLachlan, Do, and Ambroise}]{McLachlan2004AnalyzingMG}
Geoffrey~J. McLachlan, Kim-Anh Do, and Christophe Ambroise. 2004.
\newblock \href {https://api.semanticscholar.org/CorpusID:88647680} {Analyzing microarray gene expression data}.

\bibitem[{Merrill et~al.(2021)Merrill, Goldberg, Schwartz, and Smith}]{merrill2021provable}
William Merrill, Yoav Goldberg, Roy Schwartz, and Noah~A Smith. 2021.
\newblock Provable limitations of acquiring meaning from ungrounded form: What will future language models understand?
\newblock \emph{Transactions of the Association for Computational Linguistics}, 9:1047--1060.

\bibitem[{OpenAI(2023)}]{openai2023gpt4}
OpenAI. 2023.
\newblock \href {http://arxiv.org/abs/2303.08774} {Gpt-4 technical report}.

\bibitem[{Petroni et~al.(2019)Petroni, Rockt{\"a}schel, Lewis, Bakhtin, Wu, Miller, and Riedel}]{petroni2019language}
Fabio Petroni, Tim Rockt{\"a}schel, Patrick Lewis, Anton Bakhtin, Yuxiang Wu, Alexander~H Miller, and Sebastian Riedel. 2019.
\newblock Language models as knowledge bases?
\newblock \emph{arXiv preprint arXiv:1909.01066}.

\bibitem[{Pimentel et~al.(2020)Pimentel, Valvoda, Maudslay, Zmigrod, Williams, and Cotterell}]{pimentel2020information}
Tiago Pimentel, Josef Valvoda, Rowan~Hall Maudslay, Ran Zmigrod, Adina Williams, and Ryan Cotterell. 2020.
\newblock Information-theoretic probing for linguistic structure.
\newblock \emph{arXiv preprint arXiv:2004.03061}.

\bibitem[{Radford et~al.(2018)Radford, Narasimhan, Salimans, Sutskever et~al.}]{radford2018improving}
Alec Radford, Karthik Narasimhan, Tim Salimans, Ilya Sutskever, et~al. 2018.
\newblock Improving language understanding by generative pre-training.

\bibitem[{Radford et~al.(2019)Radford, Wu, Child, Luan, Amodei, and Sutskever}]{radford2019language}
Alec Radford, Jeff Wu, Rewon Child, David Luan, Dario Amodei, and Ilya Sutskever. 2019.
\newblock Language models are unsupervised multitask learners.

\bibitem[{Ravichander et~al.(2020)Ravichander, Belinkov, and Hovy}]{Ravichander2020probing}
Abhilasha Ravichander, Yonatan Belinkov, and Eduard~H. Hovy. 2020.
\newblock \href {http://arxiv.org/abs/2005.00719} {Probing the probing paradigm: Does probing accuracy entail task relevance?}
\newblock \emph{CoRR}, abs/2005.00719.

\bibitem[{Rolnick and Tegmark(2017)}]{Rolnick2017ThePO}
David Rolnick and Max Tegmark. 2017.
\newblock \href {https://api.semanticscholar.org/CorpusID:3509777} {The power of deeper networks for expressing natural functions}.
\newblock \emph{ArXiv}, abs/1705.05502.

\bibitem[{Sanh et~al.(2019)Sanh, Debut, Chaumond, and Wolf}]{sanh2019distilbert}
Victor Sanh, Lysandre Debut, Julien Chaumond, and Thomas Wolf. 2019.
\newblock Distilbert, a distilled version of bert: smaller, faster, cheaper and lighter.
\newblock \emph{arXiv preprint arXiv:1910.01108}.

\bibitem[{Shin et~al.(2020)Shin, Park, Yuk, and Lee}]{shin2020bert}
Hyeong~Jin Shin, Jeong~Yeon Park, Dae~Bum Yuk, and Jae~Sung Lee. 2020.
\newblock Bert-based spatial information extraction.
\newblock In \emph{Proceedings of the Third International Workshop on Spatial Language Understanding}, pages 10--17.

\bibitem[{Stone(1974)}]{crossval1974}
M.~Stone. 1974.
\newblock \href {https://doi.org/https://doi.org/10.1111/j.2517-6161.1974.tb00994.x} {Cross-validatory choice and assessment of statistical predictions}.
\newblock \emph{Journal of the Royal Statistical Society: Series B (Methodological)}, 36(2):111--133.

\bibitem[{Touvron et~al.(2023)Touvron, Martin, Stone, Albert, Almahairi, Babaei, Bashlykov, Batra, Bhargava, Bhosale et~al.}]{touvron2023llama}
Hugo Touvron, Louis Martin, Kevin Stone, Peter Albert, Amjad Almahairi, Yasmine Babaei, Nikolay Bashlykov, Soumya Batra, Prajjwal Bhargava, Shruti Bhosale, et~al. 2023.
\newblock Llama 2: Open foundation and fine-tuned chat models.
\newblock \emph{arXiv preprint arXiv:2307.09288}.

\bibitem[{Tucker et~al.(2021)Tucker, Qian, and Levy}]{tucker2021if}
Mycal Tucker, Peng Qian, and Roger Levy. 2021.
\newblock What if this modified that? syntactic interventions via counterfactual embeddings.
\newblock \emph{arXiv preprint arXiv:2105.14002}.

\end{thebibliography}


\end{document}